\documentclass[journal]{IEEEtran}

\IEEEoverridecommandlockouts                              

\usepackage{amsmath} 

\usepackage{amsfonts,amssymb}
\usepackage[utf8]{inputenc}
\usepackage{graphicx}
\usepackage{caption}
\usepackage{multirow}
\usepackage{booktabs}
\usepackage{hyperref}
\usepackage{subfigure}
\usepackage{bbding}
\usepackage{amsthm}
\usepackage{mathrsfs}
\usepackage[ruled,linesnumbered,lined,boxed,commentsnumbered]{algorithm2e}

\graphicspath{ {./figure/} }

\begin{document}
\title{\LARGE \bf
TrajGen: Generating Realistic and Diverse Trajectories with \\Reactive and Feasible Agent Behaviors for Autonomous Driving}

\author{${\text{Qichao Zhang}}^{1*}$, $\text{Yinfeng Gao}^{2*}$,$\text{Yikang Zhang}^{1*}$, $\text{Youtian Guo}^{1}$, $\text{Dawei Ding}^{2}$, \\ $\text{Yunpeng Wang}^{3}$,  $\text{Peng Sun}^{3}$,  $\text{Dongbin Zhao}^{1}$, ~\IEEEmembership{IEEE Fellow}$ $

\thanks{*This work is partly supported by the National Natural Science Foundation of China (NSFC) under Grants No.62173325, and by  Science and Technology Innovation 2030 'New Generation Artificial Intelligence' Major Project No. 2020AAA0103700.}
\thanks{$^{1}$ is with The State Key Laboratory of Management and Control for Complex Systems, Institute of Automation, Chinese Academy of Sciences, Beijing, 100190, China, and also with the College of Artificial Intelligence, University of Chinese Academy of Sciences, Beijing, 100049, China. (email:zhangqichao2014@ia.ac.cn, Dongbin.zhao@ia.ac.cn)}%
\thanks{$^{2}$ is with the School of Automation and Electrical Engineering, University of Science and Technology Beijing, Beijing, 100083, China.}%
\thanks{$^{3}$ is with Baidu Inc., Beijing 100085, China.}%
\thanks{* means the authors contribute equally.}
\thanks{The I-Sim simulator will be open-sourced soon.}

}

\maketitle
\markboth{IEEE TRANSACTIONS ON INTELLIGENT TRANSPORTATION SYSTEMS, VOL. , NO. , 2021
}{Roberg \MakeLowercase{\textit{et al.}}: High-Efficiency Diode and Transistor Rectifiers}

\begin{abstract}
Realistic and diverse simulation scenarios with reactive and feasible agent behaviors can be used for  validation and verification of self-driving system performance without relying on expensive and time-consuming real-world testing. Existing simulators rely on heuristic-based behavior models for background vehicles, which cannot capture the complex interactive behaviors in real-world scenarios.  To bridge the gap between simulation and the real world, we propose TrajGen, a two-stage trajectory generation framework, which can capture more realistic behaviors directly from human demonstration. In particular, TrajGen consists of the multi-modal trajectory prediction stage and the reinforcement learning based trajectory 
modification stage. In the first stage, we propose a novel auxiliary RouteLoss for the trajectory prediction model to generate multi-modal diverse trajectories in the drivable area. In the second stage, reinforcement learning is used to track the predicted trajectories while avoiding collisions, which can improve the feasibility of generated trajectories. In addition, we develop a data-driven simulator I-Sim that can be used to train reinforcement learning models in parallel based on naturalistic driving data. The vehicle model in I-Sim can guarantee that the generated trajectories by TrajGen satisfy vehicle kinematic constraints. Finally, we give comprehensive metrics to evaluate generated trajectories for simulation scenarios, which shows that TrajGen outperforms  either  trajectory  prediction  or  inverse  reinforcement learning in terms of fidelity, reactivity, feasibility, and diversity.
\end{abstract}


\section{INTRODUCTION}
Recently, self-driving has achieved widespread attention in academic and industry communities \cite{li2019deep,guo2021hierarchical}. However, how to validate and verify the performance of the decision-making and planning module is still unsolved. In industry, real-world testing with a fleet of autonomous vehicles (AVs) is a common approach. Unfortunately, such an evaluation process is both expensive and time-consuming, since valuable interactive scenarios are rare and safety-critical in the physical world. As a result, simulation has been an important evaluation tool enabling AVs' rapid development and safe deployment. 

\begin{figure}[htbp]
    \centering
    \includegraphics[width=8.5cm]{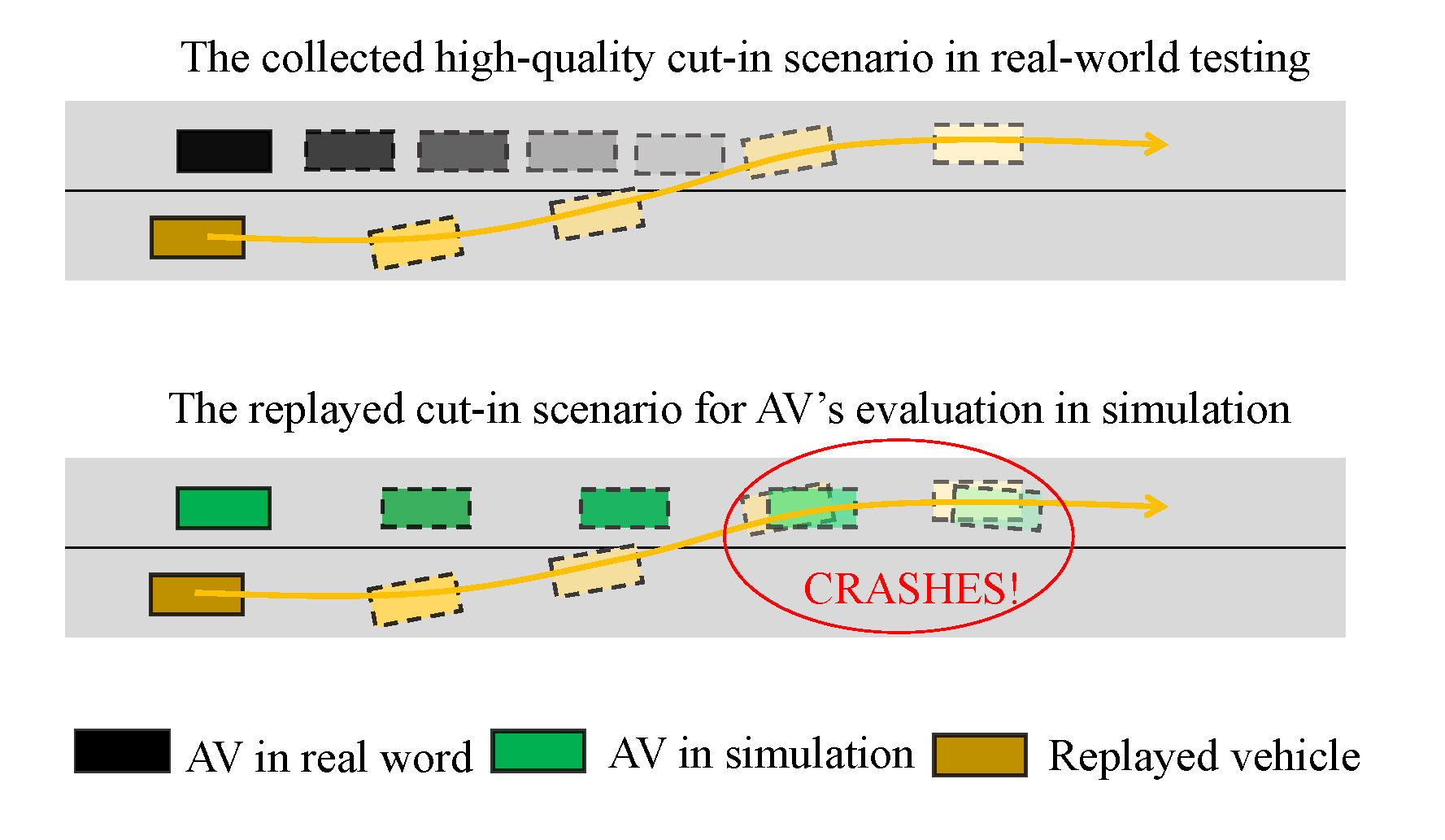}
\caption{The issue of log replay. The yellow vehicle is a replayed cut-in vehicle, and the black AV takes a yield behavior in real-world testing. However, when the green AV with an upgraded system version takes an acceleration behavior in this replayed simulation scenario, the yellow replayed vehicle will collide with the green AV. The responsibility of this unrealistic accident is due to the non-reactive behavior of the replayed vehicle, since the green AV has the path priority of starting. Note that the vehicle color fades as the timestep increases.}
\label{motivation}
\end{figure}

A common scenario generation approach in simulator environments is log replay \cite{osinski2020carla}, where the behaviors of traffic participants are replayed according to the collected log trajectories \cite{scanlon2021waymo}. However, once the behavior of AV is different from the one when the log was collected, those replayed evaluation scenarios may be unrealistic since the replayed traffic participants don't have any reactive capabilities. In other words, a lot of high-quality interactive scenario data collected by the AVs in real-world testing may be valueless to evaluate the AVs with upgraded system version based on the log replay, as shown in Fig. \ref{motivation}. The widely used reactive vehicle behavior model in existing simulators is based on the heuristic method such as Intelligent Driver Model (IDM) \cite{kesting2010enhanced}, MOBIL \cite{kesting2007general}, and so on. However, those models are difficult to model the fidelity and diversity in interactive behaviors of road testing. Feng \textit{et al.} \cite{feng2021intelligent} show that the IDM model even calibrated by real-world datasets cannot generate realistic and diverse driving trajectories for highway scenarios. 

To bridge the behavior gap between simulation and real-world agents, Wang \textit{et al.} \cite{wang2010parallel} try to generate realistic and diverse trajectories with data-driven and learning models based on real-world data. The common approaches are trajectory prediction or motion forecasting \cite{li2020prediction,zhao2020spatial} and reinforcement learning (RL) related methods \cite{bhattacharyya2019simulating,zheng2021objective}.  
For the trajectory prediction model, the stochastic multi-modality behaviors can be captured based on the real-world data \cite{suo2021trafficsim,bergamini2021simnet}. However, the feasibility of generated trajectories cannot be guaranteed such as avoiding the unrealistic collision and satisfying vehicle kinematic constraints, as the agent is considered as a particle model. 
On the other hand, for the RL related methods \cite{chen2021interpretable, li2021Harmonious}, the collision rate can be reduced by adding the collision penalty. Considering the vehicle model in the simulator, the vehicle kinematic constraints can be also satisfied. Unfortunately, the generated trajectories lack diversity since the nearly-optimal policy is trained to maximize the cost function, which is difficult to obtain diverse human-like behaviors \cite{hang2020human}.

\begin{figure*}[htbp]
\centering
\includegraphics[width=18cm]{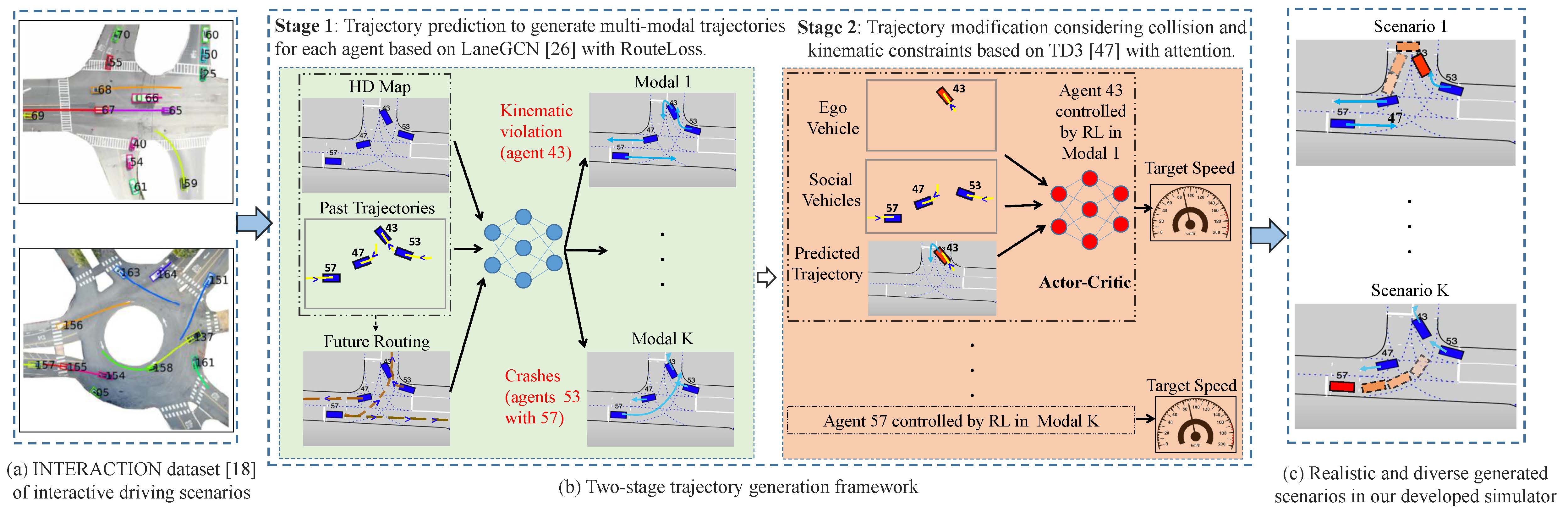}
\caption{The proposed TrajGen with a two-stage trajectory generation framework. The blue vehicles are controlled by the trajectory prediction model. As shown in sub-figure(b), the vehicles with kinematic violation (agent 43) or collisions (agent 57) in stage 1 are controlled by the RL model to modify their predicted trajectories in stage 2. For the proposed TrajGen, the fidelity and diversity of trajectories are provided by the trajectory prediction, and the reactivity and feasibility of trajectories are improved by RL. Finally, some realistic and diverse scenarios can be generated in I-Sim as shown in sub-figure(c).}
\label{TrajGen}
\end{figure*}

Motivated by those, we present TrajGen to generate realistic and diverse scenarios with reactive and feasible agent behaviors based on naturalistic driving data, which is a two-stage trajectory generation framework shown in Fig. \ref{TrajGen}. The main contributions of this paper are as follows.

\begin{itemize}
\item A novel two-stage trajectory generation framework is proposed for simulation scenarios, by combining the trajectory prediction model to guarantee the fidelity and diversity of trajectories, and the RL model to improve the reactivity and feasibility;

\item To improve the performance of trajectory prediction, an auxiliary RouteLoss is proposed by utilizing the prior knowledge of High-Definition Map (HD Map). In addition, the social attention mechanism is introduced in the RL model to better capture interaction features between the ego vehicle and nearby vehicles.

\item We develop a data-driven simulator I-Sim with an  OpenAI  Gym  environment based on the INTERACTION dataset \cite{zhan2019interaction}, which can be used for closed-loop RL training and evaluation based on the real-world dataset. Experimental results in I-Sim show that TrajGen outperforms either trajectory prediction or inverse RL in terms of comprehensive  metrics.

\end{itemize}

\section{Related Work}

In this section, we will discuss relevant trajectory generation methods about multi-modal trajectory prediction, RL based driving behaviors, and existing  driving simulators. 

\subsection{Multi-modal Trajectory Prediction}
Trajectory prediction is used to predict the agent's future trajectories based on given past trajectory and context information. Recently, deep learning based methods have been developed to capture the multi-modal trajectories, which can be divided into two categories: anchor-based and anchor-free prediction. The anchor-based prediction can be considered as conditional forecasting, where the agent's prediction model is conditioned on different types of anchors. \cite{chai2020multipath} and \cite{song2021learning} condition the agent's future trajectories on a pre-defined set of anchor trajectories, while \cite{zhao2020tnt} and \cite{rhinehart2019precog} propose goal-conditional multi-modal trajectory prediction based on sparse anchor goals with heuristic goal selection algorithms. \cite {narayanan2021divide} proposes the trajectory prediction with lane anchors, which uses existing lane centerlines as anchors. For anchor-free prediction, \cite{tang2019multiple} proposes a sequential probabilistic latent variable generative model to generate multi-modal interactive trajectories jointly for variable agents. \cite{gao2020vectornet} introduces an efficient vectorized representation with graph attention networks for multi-modal prediction. To capture the complex interactions between agents and maps, LaneGCN \cite{liang2020learning} is proposed with an interaction fusion network, which exploits HD Map with vectorized representation explicitly.

Recently, TrafficSim \cite{suo2021trafficsim} uses a scene-consistent trajectory predictor as the multi-agent behavior model to simulate realistic and diverse traffic scenarios. To avoid generating collision trajectories, a common-sense loss is designed. Similarly, SimNet \cite{bergamini2021simnet} uses a multi-modal trajectory predictor as the agent reactive behavior model to generate trajectories for background vehicles. RouteGAN \cite{yin2021diverse} can generate diverse interaction behaviors by controlling the agents separately with desired styles and given final goals. Although the diversity and reactivity are improved based on trajectory prediction models, the feasibility of those produced unconstrained trajectories, such as the collision-free behaviors and vehicle kinematic feasibility, can not be guaranteed. However, the feasibility of generated trajectory is essential for realistic simulation scenarios. 

\subsection{RL based Driving Behaviors}

Inverse RL is widely used to simulate driving behaviors from the demonstration, which aims to infer the reward function  and learn the control policy \cite{chen2021IRLDriving} .  Generative adversarial imitation learning (GAIL) \cite{ho2016generative} is used to simulate multi-agent driving behaviors with the parameter sharing technique in \cite{bhattacharyya2018multi}. Considering the complexity of the real-world traffic, \cite{bhattacharyya2019simulating} proposes a reward augmented imitation learning (RAIL) to learn human driving behavior by embedding the prior knowledge in reward augmentation. \cite{zheng2021objective} uses the adversarial inverse RL (AIRL) to train traffic simulating the agent's car-following behavior.  \cite{o2018scalable} utilizes the model-based GAIL to generate rare-event scenarios for vision-based end-to-end autonomous driving algorithms. However, these methods are difficult to capture the multi-modal properties of the traffic since inverse RL aims to obtain a nearly-optimal policy by maximizing the learned cost function. 

Recently, some researchers try to apply RL \cite{zhao2017model} in generating diverse simulation scenarios. \cite{chen2021adversarial} uses deep deterministic policy gradient without exploration noises to generate diverse lane-change scenarios. \cite{shiroshita2020behaviorally} designs a distinct policy set selector to balance diversity and driving skills. However, Without relying on real-world data, realistic behaviors are difficult to be generated for complex interactive scenarios. The comparison between existing methods and our proposed TrajGen is shown in Table \ref{COMPARISON RESULTS}.

\begin{table}[h]
\caption{Comparison of related methods}
\label{COMPARISON RESULTS}
\setlength\tabcolsep{2pt}
\begin{center}
\begin{tabular}{c|c c c c}
\hline 
Methods & Fidelity  & Reactivity & Feasibility & Diversity\\
\hline 
Log replay\cite{osinski2020carla} & \Checkmark &\Checkmark &  \XSolid &  \Checkmark \\
\hline
Heuristic method\cite{kesting2010enhanced,kesting2007general} & \XSolid
 &\Checkmark & \Checkmark  &  \XSolid \\
\hline
Trajectory prediction\cite{suo2021trafficsim, liang2020learning} & \Checkmark &  \Checkmark &\XSolid &  \Checkmark \\
\hline
RL\cite{chen2021adversarial, shiroshita2020behaviorally} & \XSolid &\Checkmark &  \Checkmark & \Checkmark \\
\hline
TrajGen (ours) & \Checkmark &\Checkmark &  \Checkmark &  \Checkmark\\
\hline 
\end{tabular}
\end{center}
\end{table}

\subsection{Existing Driving Simulator}
The autonomous driving simulator is essential for RL based policy training, as well as the closed-loop evaluation. Recently, researchers have leveraged the racing game TORCS \cite{li2019reinforcement}, microscopic traffic simulator SUMO \cite{lopez2018microscopic}, large-scale city-level traffic simulator CityFlow \cite{tang2019cityflow}, and hand-coded simulation environment Highway-env \cite{highway-env} to train and evaluate RL based driving policy. In particular, the provided maps and behavior models for background vehicles in these simulators are simplistic, and can not reflect the complex multi-agent interaction in the real world. SUMMIT \cite{luo2020simulating} and CARLA \cite{dosovitskiy2017carla} are open high-fidelity simulators with realistic visual appearance. Different from CARLA, SUMMIT provides many real-world maps, and the agent behavior model is based on a deterministic trajectory prediction model considering the physical constraints.  BARK \cite{bernhard2020bark} provides a behavior benchmark with the heuristic model, RL model, and dataset tracking model with a few scenarios from the INTERACTION dataset \cite{zhan2019interaction}. 
The newly developed simulator SMARTS \cite{zhou2020smarts} and MetaDrive \cite{li2021metadrive} provide excellent interaction environments for multi-agent RL agents in atomic traffic scenes. Different from SMARTS and MetaDrive, the HD Maps and vehicle behaviors in I-Sim are completely based on real-world complex interactive scenarios to guarantee the fidelity, which can facilitate the generalizable RL research in  autonomous driving.

\section{Generating  Realistic  and  Diverse  Scenarios}

\subsection{Overview}

The simulation scenarios are usually generated as follows:
1) scenarios initialization: specifying the initial scene layout with the road topology and agents' initial states such as position and velocity. 2) forward simulation: simulating the motion of dynamic agents forward, and 3) scenarios termination: ending the scenario when the designed termination conditions are met. In this paper, the initial scene is provided by the INTERACTION dataset shown in Fig. \ref{TrajGen}(a). And the termination conditions are unrolling for 8\textit{s} during forward simulation, or all agents reach their log trajectories' endpoints. We focus on the motion of dynamic agents for forward simulation by using TrajGen.

Given an initial scene with HD Map $M$ and historical  joint states of $N$ dynamic agents, our goal is to design the agent behavior trajectories for their forward simulation. The process of forward simulation can be described as a sequence of joint state  ${h}_t=\{z_t^1, z_t^2,...,z_t^N\}$ with the unrolling horizon $T$, where $z_t^n$ means the state of the $n$th agent including its 2D position and speed information at time $t$. Given a sequence of historical joint states derived from the perception system, the multi-modal future trajectories should be captured for each agent. For example, agent 57 in Fig. \ref{TrajGen}(b) can go straight in modal 1 or turn left in modal $K$. In addition, its behavior should satisfy the vehicle kinematic constraint and the common sense such as avoiding collisions. In the following, we describe the two-stage trajectory generation framework combing trajectory prediction and RL tracking, which can generate realistic and diverse trajectories with reactive and feasible behaviors. 

\subsection{Stage 1: Trajectory prediction to generate multi-modal trajectories}
For the  multi-modal trajectory prediction stage, we aim to generate a joint distribution $P(Y_{pred}|Y_{hist},M)$ of multi-modal trajectories $Y_{pred}=\{h_{t+1},h_{t+2},...,h_{t+T_{pred}}\}$ for $N$ agents in parallel with a sequence of historical joint states $Y_{hist}=\{h_{t-T_{hist}+1},h_{t-T_{hist}+2},...,h_t\}$, and HD Map context $M$. $T_{pred}$ denotes the total prediction steps during the unrolling horizon $T$, and $T_{hist}$ means the historical steps. 
To capture the complex and diverse interaction behaviors, we employ the LaneGCN model described in \cite{liang2020learning} with vectorized map representation on the INTERACTION dataset. A FusionNet with multiple interaction blocks stacked provided in \cite{liang2020learning} is used to extract interaction features. Based on the extracted features, multiple prediction headers can predict multi-modal trajectories with the following max-margin classification loss and  smooth L1 regression loss. 
\begin{equation}
        L_{cls} = \frac{1}{N(K-1)} \sum_{n=1}^N \sum_{k \neq \hat{k}} \max(0, c^{n,k} + \epsilon - c^{n,\hat{k}})
    \end{equation}
    \begin{equation}
        L_{reg} = \frac{1}{NT} \sum_{n=1}^N \sum_{t=1}^{T_{pred}} reg(z_{t}^{n,\hat{k}} - z_{t}^{n,*})
    \end{equation}
where $\hat{k}$ means the best modal with \textit{Minimum Final Displacement Error} (minFDE), $c^{n,k}$ means the confidence score of the $k$th modal trajectory for the $n$th agent, $\epsilon$ is the margin, and $z^*$ means ground truth trajectory in prediction horizon. There are $N$ agents and $K$ modals.

\begin{figure}[htbp]
    \centering
    \includegraphics[width=8.5cm]{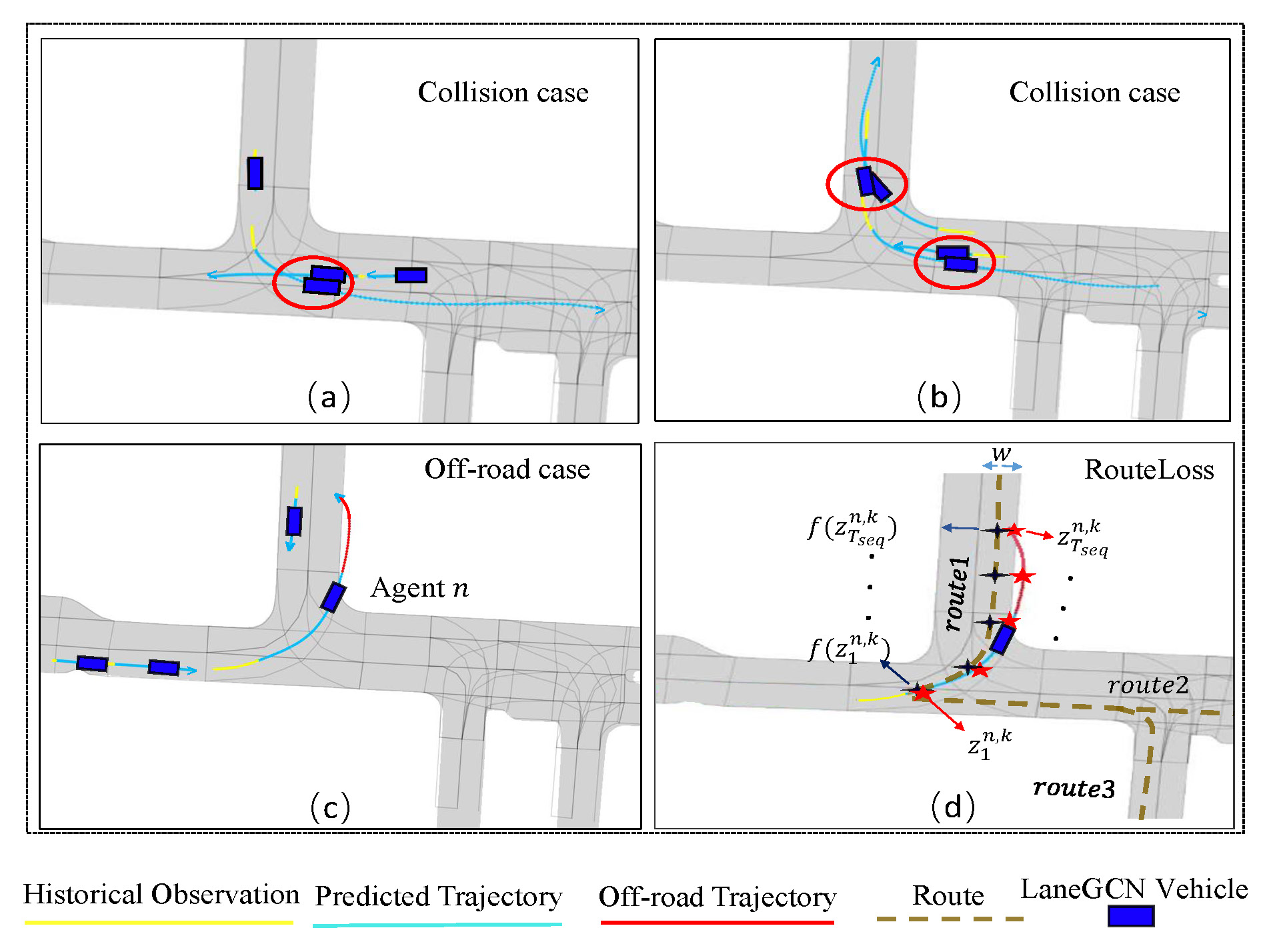}
\caption{Illustration of the collision (a,b) and off-road (c) behaviors generated by the LaneGCN model in INTERACTION dataset. To penalize the off-road behaviors, an auxiliary RouteLoss (d) is designed in stage 1, where the red stars denote the sampled waypoints $z_j^{n,k}$ from the $k$th predicted trajectory of agent $n$, and black diamonds denote the projected route points $f(z_{j}^{n,k})$ from the nearest route. The collision behaviors are addressed in stage 2.}
\label{LaneGCNPromblem}
\end{figure}

INTERACTION is a dataset that provides a fixed Bird-Eye-View (BEV) above complex traffic scenes including intersections, roundabouts, and merging roads. For the INTERACTION dataset, we find that the trajectories generated by LaneGCN shown in Fig. \ref{LaneGCNPromblem}(a-c) have a high probability of off-road and collision. To utilize  the prior knowledge of HD Map, \cite{niedoba2019improving} propose an auxiliary off-road loss to penalize off-road trajectories. In \cite{anonymous2021improving}, an auxiliary LaneLoss is introduced to encourage diverse trajectories enough to cover every possible maneuver. Note that LaneLoss is designed based on the normal distance of the endpoint of the predicted trajectory from its nearest lane waypoint in the Frenet-Serret Frame. Obviously, this LaneLoss is not helpful to improve the  quality of off-road trajectory with an endpoint in the lane like \ref{LaneGCNPromblem}(c). Motivated by this, we design a novel auxiliary RouteLoss as follows:
\begin{equation}
 L_{route} =\sum_{n=1}^N \sum_{k=1}^K \min_{f \in L} \sum_{j \in Tseq} \max(0, d(z_{j}^{n,k}, f(z_{j}^{n,k})) - m)
\end{equation}
where $L$ contains all available routes according to agents' history and the HD Map $M$ using a deep limited depth first search (DFS) algorithm over the map graph. $Tseq$ is the set of sampled timestamps from the predicted trajectory including startpoint and endpoint. The deviation $d(\cdot)$ represents the L2 distance between the $n$th agent's prediction trajectory waypoint $z_{j}^{n,k}$ for modal $k$ at timestamp $j$ and its corresponding projected route point $f(z_{j}^{n,k})$ on the nearest route, as shown in Fig. \ref{LaneGCNPromblem}(d). The lane width $m$ is used to avoid forcing agents to stay exactly in the lane. RouteLoss is designed to make the generated trajectory close to the nearest available route, which is helpful to avoid the off-road trajectory without sacrificing diversity.  

So the total loss of trajectory prediction is 
\begin{equation}
  L_{total} = L_{cls}+\alpha L_{reg} + \beta L_{route}
  \label{eq4}
\end{equation}
where $L_{cls}$ is the max-margin loss and $L_{reg}$ is smooth L1 loss defined in Eq.(1) and Eq.(2),  $\alpha$ and $\beta$ are normalization parameters. To avoid the collisions like Fig. \ref{LaneGCNPromblem}(a,b), a common-sense auxiliary loss provided in \cite{suo2021trafficsim} is used in \cite{anonymous2021urban}, the experimental results show that the acceleration over the scope phenomenon happens because the vehicle kinematic constraints are not taken into account. We address this challenge by using trajectory modification based on RL and considering the vehicle kinematic model in the simulator.

\subsection{Stage 2: Trajectory modification to avoid collisions with RL}
RL aims to improve the quality of the predicted trajectories generated in stage 1, especially for collision trajectories. It can be considered as the collision avoidance and trajectory tracking task. Considering the continuous action space, the specific RL algorithm used in this paper is Twin Delayed Deep Deterministic Policy Gradient (TD3) \cite{fujimoto2018addressing} with the social attention mechanism \cite{leurent2019social}.

\subsubsection{RL method}
\   

\textbf{Definition:} In stage 2, RL agents interact with the environment and learn to track prediction trajectories generated by LaneGCN while avoiding collisions by maximizing their cumulative reward. Here we define such tasks as Markov Decision Processes (MDP), which can be represented by the tuple $(\mathcal{S}, \mathcal{A}, \mathcal{P}, \mathcal{R}, \gamma)$, where
$\mathcal{S}$ is the state space, $\mathcal{A}$ is the action space, $\mathcal{P}: \mathcal{S} \times \mathcal{A} \times \mathcal{S} \to [0, 1]$ is the transition function which maps the state $s_t$ and action $a_t$ to a probability distribution of the next states $s_{t+1}$, $\mathcal{R}: \mathcal{S} \times \mathcal{A} \to \mathbb{R} $ is the reward function which maps the state and its corresponding action to a scalar reward value, and $\gamma \in (0, 1)$ is the discount factor. The goal of agents is to find a policy $\pi : \mathcal{S} \to \mathcal{A}$ which maximizes the expected discount cumulative reward, i.e., return:
\begin{equation}
  \max E[\sum_{t=0}^{\infty} \gamma^{t} R(s_t, \pi(s_t)) ]
\end{equation}
We also define the value function $V^\pi: \mathcal{S} \to \mathbb{R}$, which maps the state to the expected return with running policy $\pi$, and the Q-value function $Q^\pi: \mathcal{S} \times \mathcal{A} \to \mathbb{R}$, which maps the state-action pair to the expected return with running $\pi$.

\textbf{TD3 algorithm:} TD3 \cite{fujimoto2018addressing} is a popular deep RL method with the Actor-Critic architecture, where both actor and critics are represented using deep networks. In TD3, the actor $\mu$ is a policy network parameterized by $\theta$, which takes state as input and outputs an action, the critic $Q$ are composed of two Q-value approximation networks parameterized by $\phi_1$ and $\phi_2$, which take state and its corresponding action as input and output the Q-value. We use $\theta'$, $\phi_1'$, and $\phi_2'$ to represent the parameters of the target networks of actor and critics. As an off-policy method, TD3 needs a replay buffer $\mathcal{B}$ to store experiences. Every time a batch of experiences are sampled from the replay buffer, and the critic will be optimized based on the loss $L(\phi_i),i=1,2$: 
\begin{equation}
    L(\phi_i) = E_{(s_t, a_t, r_t, s_{t+1}) \sim \mathcal{B}} [q(s_t, a_t)-Q_{\phi_i}(s_t,a_t)]^{2}
\end{equation}
where $q(s_t, a_t)$ is the target value:
\begin{equation}
    q(s_t, a_t) = r_t + \gamma \min_{i=1,2} Q_{\phi_i'}(s_{t+1}, \tilde{a})
\end{equation}
with $\tilde{a} = \mu_{\theta'}(s_{t+1})+\delta, \delta \sim$ clip$(\mathcal{N}(0,\sigma), -l, l)$ the target smoothing action, $\delta$ is a random noise clipped by $\pm l$.

After updating $\phi_1$ and $\phi_2$, the actor's parameter $\theta$ is updated by the deterministic policy gradient:
\begin{equation}
    \nabla_\theta J(\theta) = E_{s \sim \mathcal{B}} [\nabla_a Q_{\phi_1}(s,a) \lvert_{a=\mu_\theta (s)} \nabla_\theta \mu_\theta (s)]
\end{equation}
Note that the actor network is updated once after updating a fixed number of $d$ times of the critic networks in TD3 due to the “delayed policy updates” setting. 

The parameters of the target networks, $\phi_1'$, $\phi_2'$ and $\theta'$, are updated softly to ensure the learning process is stable:
\begin{equation}
    \phi_i' = \tau \phi_i + (1-\tau) \phi_i'
\end{equation}
\begin{equation}
    \theta' = \tau \theta + (1-\tau) \theta'
\end{equation}
with $\tau \ll 1$. $\tau$ is a normalized hyper-parameter that controls the change speed of the target networks.

\textbf{Social attention mechanism:} Attention mechanism is useful to find the internal dependencies between different input instances for deep networks, which has been proved effective in many research fields. To help RL agents better capture dependencies between the ego vehicle and nearby vehicles while making a decision, social attention mechanism  \cite{leurent2019social} is used in our actor network. Note that the ego vehicle means the agent controlled by the RL model.

The social attention mechanism function used in this paper is a soft attention mechanism function similar to \cite{2017Attention}. Assume there are $M$ social vehicles that are near to the ego vehicle, the attention values which are output by the social attention mechanism function can be defined as:
\begin{equation}
    Attention(Q^A,K^A,V^A) = softmax (\frac{Q^A{K^A}^T}{\sqrt{d_k}})V^A
\end{equation}
where $Q^A = [q_e]$ is a single query calculated by processing the ego vehicle’s features with a nonlinear projection $P_q$, $K^A = [k_e, k_1, ..., k_M]$ are keys calculated by processing the ego and nearby vehicles' features with another nonlinear projection $P_k$. Through the dot product between $Q^A$ and $K^A$, the similarities between the query and the keys are generated, which are then scaled by $1/\sqrt{d_k}$. $d_k$ is the dimension of $q_e$ and $k_j, j=e,1,...,M$.
The output of this softmax function is defined as $attention \ weights$, which indicates ego's attention to nearby social vehicles including itself. Finally, the dot product between the $attention \ weights$ and values $V^A = [v_e, v_1, ..., v_M]$, which are calculated by processing the ego and nearby vehicles' features with the nonlinear projection $P_v$, is the result of the social attention mechanism function.
Note that the nonlinear projections $P_q$, $P_k$ and $P_v$ are all realized by multi-layer perceptron in this paper. 

\subsubsection{RL settings}
\   

\textbf{Observation:} A 48-dimension vector is used to represent RL controlled ego vehicle's current observation, which can be written as $o =[o_r, o_e, o_1, ...o_5]$. As shown in Table \ref{OBSERVATION REPRESENTATIONS}, the observation vector $o$ can be divided into three major parts: $o_r$ contains ego vehicle's target position, target speed value, and route trend. Specifically, we obtain the ego vehicle's relative position and relative speed according to the predicted trajectory as its target position and target speed. The route trend consists of heading errors between the ego vehicle and 5 closest forward prediction waypoints. $o_e$ refers to ego vehicle's shape and its movement trend, specifically, the length and width of the ego vehicle, ego vehicle's current speed value and its position in the next second if it remains current speed and heading. $o_1,...o_5$ describe the observation of 5 social vehicles, each of them contains the length, width, coordinate, current speed value, and current heading's cosine and sine value of a social vehicle.

Note that all position and heading observations mentioned above are relative to the ego's coordinate system. We design a simple filter to select social vehicles, whose rules can be described below:


\begin{align*}
\begin{split}
\left \{
\begin{array}{lr}
    d_{ei} < 30m \\
    y_{e_{i}} > -12m \\
\end{array}
\right.
\end{split}
\end{align*}
where $d_{ei}$ indicates the distance between ego vehicle and a nearby vehicle $i$, $y_{e_{i}}$ indicates a nearby vehicle $i$'s Y-axis coordinate under the ego's coordinate system. Only the closest 5 vehicles are considered as social vehicles if there are more than 5 vehicles that are kept through the filter.

\begin{table}[h]
\caption{Observation Representations}
\label{OBSERVATION REPRESENTATIONS}
\begin{center}
\begin{tabular}{c|c|c}
\hline
Component & Description & Length \\
\hline
$o_r$ & ego's target posture and route trend & 1*8  \\
$o_e$ & ego's shape and movement trend & 1*5 \\
$o_i (i=1,2,...,5)$ & social vehicles' shape and posture  & 5*7 \\
\hline
$o$ & complete observation vector & 48 \\
\hline
\end{tabular}
\end{center}
\end{table}
\textbf{Action:} 
Since most collisions are caused by unreasonable longitudinal behavior, we consider the ego vehicle's longitudinal target speed as the action space. Specifically, the output of the policy network is a normalized scalar, and it is linearly mapped to $(0, 10)m/s$ as the longitudinal target speed. Then, a longitudinal PID controller $\rm{PID}_{acc}$ is applied to control the acceleration based on the output value. For the lateral movement, another PID controller $\rm{PID}_{angle}$ is designed to control the front wheel angle to track the trajectories given by stage 1. 


\textbf{Reward:} 
The main purpose of using RL is to avoid collision while track the predicted trajectories. Thus, the reward function $r$ is designed as $r=r_{collision} + r_{tracking} + r_{step}$ with
\begin{align*}
\begin{split}
\left \{
\begin{array}{lr}
    r_{collision} = -500*(1+v\_norm) \\
    r_{tracking} = 1-0.2*d_{ep} \\
    r_{step} = -0.5
\end{array}
\right.
\end{split}
\end{align*}
where $v\_norm$ is the ego vehicle's normalized speed value. It is easier for the ego vehicle to learn to avoid active collisions through dynamic adjustment $r_{collision}$ with $v\_norm$. 
We use $r_{tracking}$ to reduce the distance between ego vehicle and its tracking trajectory, where $d_{ep}$ refers to ego vehicle's distance to its current predicted position. $r_{step}$ is used to prevent RL from learning a negative strategy such as staying at the startpoint.

Combining the trajectory prediction in stage 1 and trajectory modification in stage 2, the pseudo-code of the proposed TrajGen is given in Algorithm 1.

\begin{algorithm}
  \caption{Algorithm for TrajGen}
  \label{alg1}
  \SetKwInOut{Input}{input}
  \SetKwInOut{Output}{output}
  
 \Input {INTERACTION dataset with trajectory data and HD Map.}
  \Output {Multi-modal trajectories for $N$ agents.}

  \textbf{Stage 1: Train trajectory prediction model}

   Randomly initialize LaneGCN network, and the hyper-parameters $K$, $\alpha$, $\beta$. 
   
   Search each agent's routes using depth limited DFS over the map graph.\\
   \For {iterations $iter=0 \ \KwTo \ ITER$} {
   \For {modes $k=1 \ \KwTo \  K$ }{
   \For {agent $n=1 \ \KwTo \  N$ }{
    output the confidence score $c^{n,k}$ and the predicted trajectory $z^{n,k}$.
    }
    }
  Compute the total loss described in Eq.(\ref{eq4}).
  
  Update all parameters via ADAM optimizer.
  }
  \BlankLine
  
  \textbf{Stage 2: Train trajectory modification RL model}
  
  Initialize critic's networks $Q_{\phi_1}$, $Q_{\phi_2}$ and social attentive actor's network $\mu_\theta$ with random parameters $\phi_1$, $\phi_2$, $\theta$ for TD3 algorithm.
 
  Initialize target networks $\phi_1' \gets \ \phi_1$, $\phi_2' \gets \  \phi_2$, $\theta' \gets \  \theta$.
  
  Initialize replay buffer $\mathcal{B}$.
  
  \For{iterations $iter=1 \ \KwTo \ ITER$}{
   Select action with exploration noise $a = \mu(s)+\delta, \delta \sim \mathcal{N}(0,\sigma)$.
   
   Observe reward $r$ and next state $s'$.
   
   Store transition tuple $(s,a,r,s')$ in $\mathcal{B}$.
   \\
   
   Sample a mini-batch of transitions $(s,a,r,s')$ from $\mathcal{B}$.
   
   Update critic's parameters $\phi_1$ and $\phi_2$ by minimizing $L(\phi_i)$ described in Eq.(6).
   
   \If{$t \ {\rm mod} \ d$}{
   Update actor's parameters $\theta$ with Eq.(8).
   
   Update target networks' parameters $\phi_1'$, $\phi_2'$ and $\theta'$ with Eq.(9) and Eq.(10).
   }
   }
\end{algorithm}

\subsection{Simulator: I-Sim}
\begin{figure}[htbp]
    \centering
    \includegraphics[width=8.5cm]{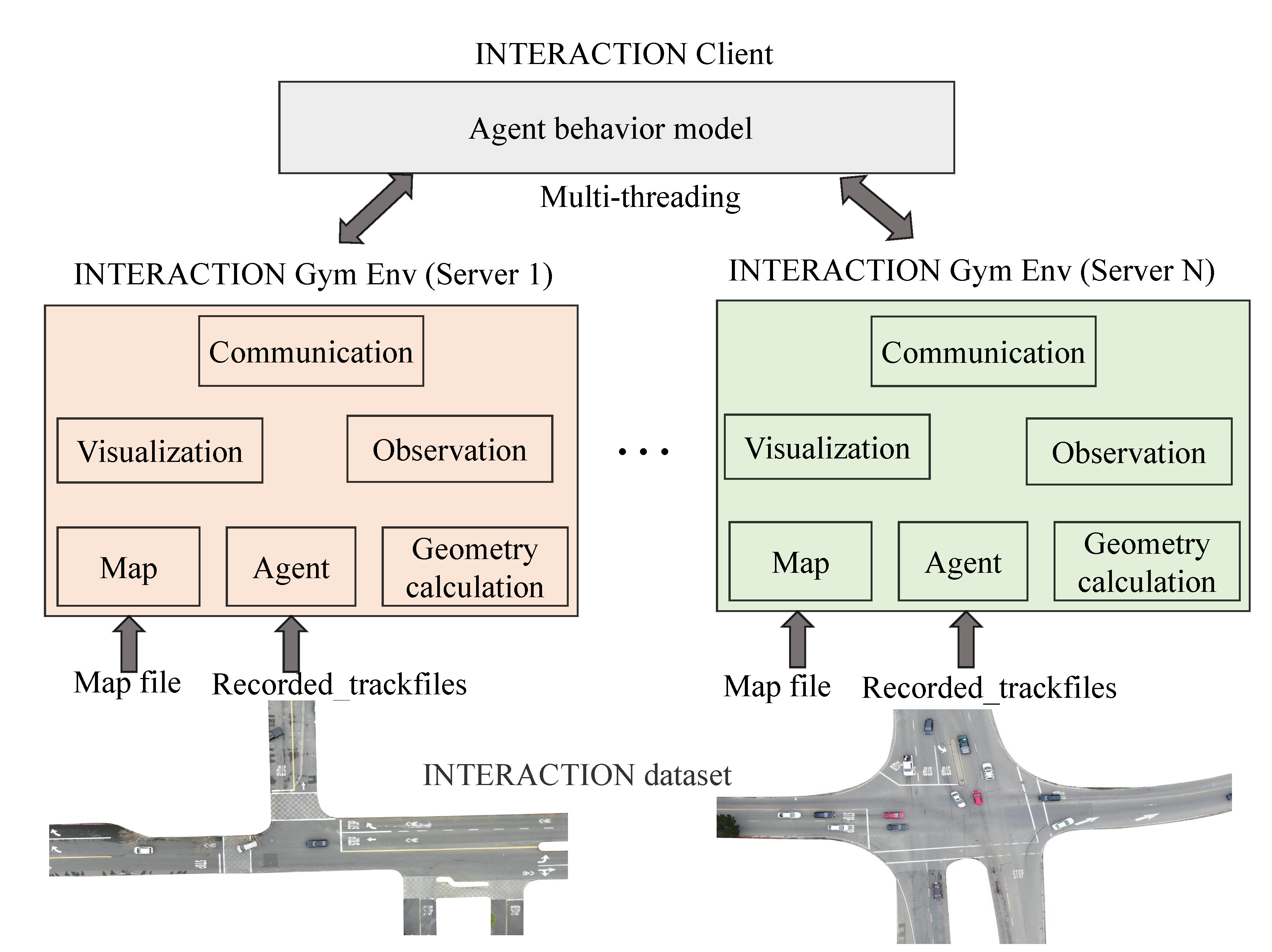}
\caption{An overview of I-Sim that simulates massive mixed traffic based on INTERACTION dataset.}
\label{Interaction}
\end{figure}

To train the RL agent, we develop I-Sim, a data-driven simulator with an OpenAI Gym environment based on the INTERACTION dataset. The overview of I-Sim with a three-layer architecture is shown in Fig. \ref{Interaction}, which can support effective RL and inverse RL training in parallel. 

1) The bottom layer can load centimeter-accurate HD Map files and recorded\_trackfiles for agents from the INTERACTION dataset. Based on this layer, we can obtain the replayed scenarios for the INTERACTION dataset. In addition, the bottom layer is responsible for providing the agent's kinematic model and geometry calculation functions such as the distance and angle calculation function between two rectangles. The kinematic model can be considered as the transition modeling for the traffic vehicles in MDP. In I-Sim, we use a bicycle model \cite{polack2017kinematic} for agents given by 

\begin{equation}
\left
\{\begin{array}{l}
x_{t+1}=x_{t}+v_{t} \cos \left(\psi_{t}+\omega\right) \times \Delta t \\
y_{t+1}=y_{t}+v_{t} \sin \left(\psi_{t}+\omega\right) \times \Delta t \\
\psi_{t+1}=\psi_{t}+\frac{v_{t}}{l_{r}} \sin (\omega) \times \Delta t \\
v_{t+1}=v_{t}+a c c \times \Delta t
\end{array}
\label{k-model}
\right.
\end{equation}

with $\omega = \tan^{-1}(\frac{l_r}{l_r+l_f} \tan{(\delta_f)})$, where $t$ and $\Delta t$ indicate current timestamp and delta time separately, $x$ and $y$ represent coordinates, $\psi$ represents vehicle's yaw angle, and $v$ represents vehicle's speed. The vechcle is controlled through the acceleration $acc$ and the front wheel angle $\delta_f$. Note that $l_f$ and $l_r$ are distances from the front and rear wheels to the vehicle center of gravity. To get $l_f$ and $l_r$ of vehicles with different shapes for simplicity, we assume that the wheelbase length is a fixed ratio of 0.6 of the vehicle's length and the gravity center of the vehicle is in its middle. To guarantee the feasibility, the acceleration range is set to $[-3, 3]m/s^2$, and the max front wheel angle is set to $\pm30$ degrees according to the INTERACTION dataset.

2) The middle layer is designed to give observation modeling and visualization tools. In I-Sim, we provide two kinds of common observation representations, which are vector observation and BEV observation. Note that the vector observation is used in the proposed TrajGen. The visualization tools can be used to analyze the agent behaviors and trajectory quality qualitatively.

3) The top layer is used for communication with the server-client pattern, which can carry out parallel simulation for efficient RL training. In detail, the client can obtain required data from multiple servers in parallel. After the RL policy is updated, the corresponding output action for each server is transmitted to its controlled agent. Based on the bicycle model in Eq.(\ref{k-model}), the next states can be obtained. Meanwhile, it also supports the scenario generation ability by providing multiple control APIs, which allows us to obtain observations and generate trajectories for multiple vehicles based on the proposed method simultaneously in one single server thread. 
 



\section{Experiments}

\subsection{Dataset} 

We use the INTERACTION dataset \cite{zhan2019interaction}, which contains different highly interactive and complex urban scenarios from different countries. 
{To validate the trajectory prediction performance with the proposed auxiliary RouteLoss in stage 1}, we first compare the performances on the INTERACTION dataset with intersection scenarios for $T=3s$ prediction horizon, which is a common setting for the trajectory prediction task. Note that the trajectories are sampled at 10Hz with (0, 2]\textit{s} for observation and (2, 5]\textit{s} for future prediction.
Since the unrolling horizon is 8\textit{s} for our trajectory generation task, we additionally extract 11938 snippets with 10\textit{s} length, where 8437 are set for training, 2365 for validation, and 1136 for testing. The prediction model performances are also validated for the $T=8s$ prediction horizon.

{To evaluate the trajectory quality generated by TrajGen in highly interactive scenarios}, we extract 425 snippets containing safety-critical situations from the USA$\_$Intersection$\_$EP0 in INTERACTION dataset as the validation dataset. Firstly, we deploy the trained LaneGCN with RouteLoss predictor on this safety-critical validation set with 10\textit{s} length for RL training in stage 2. During the RL training process, we use the HD Map and (0,2]\textit{s} trajectories as the scenario initialization, and the subsequent $T=8s$ for forward simulation in the developed I-Sim. After training, the converged RL model is used to modify those trajectories generated by the predictor, which can reduce collisions and satisfy the kinematic constraints.

\subsection{Metrics}
Evaluating  behavior simulation is still challenging since there is
no singular quantitative metric that can fully capture the quality of the
generated simulation scenarios. 
In this paper, we design the metrics in four aspects as shown in Table I, which are all important to evaluate the trajectory quality for the simulation scenarios.
\subsubsection{Fidelity}
We use distance-based scenario
reconstruction metrics to evaluate the  fidelity between generated trajectories with the ground truth. The fidelity of generated trajectories should be a primary concern.

For the prediction task in stage 1, since the time horizons of generated trajectories are the same, we adopt the widely used  minADE/minFDE by selecting the best matching prediction modal, and meanADE/meanFDE by averaging over the set of $K=6$ predicted trajectories. In addition, we also provide the off-road rate (OR) as follows.

\[
\rm{OR} = \frac {\textit{Num}( \rm{off\mbox{-}road\   trajectories  )}} {\textit{Num}( \rm{all\ generated \  trajectories)}}
\]
where $\textit{Num}(\cdot)$ means the number.

For simulation scenarios evaluation, since the time horizons of trajectories generated by TrajGen may be different for different agents, we use the root mean square error (RMSE)  of position \cite{zheng2021objective} to evaluate the mismatch between the generated trajectories with the ground truth.
\begin{align}
 {\rm{RMSE}} = \frac{1}{N} \sum_{n=1}^N \sqrt{\frac{1}{T_{pred}} \sum_{t=1}^{T_{pred}} (z_{t}^n-\hat {z}_t^n)^2}
\end{align}

\subsubsection{Reactivity} Motivated by \cite{bergamini2021simnet}, we measure the agent's reactivity based on the collision rate in synthetic scenarios, where a static car is placed in front of a moving car controlled by the behavior model. This requires the trailing car to react by stopping. 233 scenarios are chosen from the safety-critical validation set for reactivity testing. We report the synthetic collision rate (SCR) as follows.

\[
\rm{SCR} = \frac {\textit{Num}(\rm{ collision \ scenarios)}} {\textit{Num}(\rm{all\ synthetic \  scenarios)}}
\]

\subsubsection{Feasibility}
We consider the feasibility of generated trajectories from the following two aspects.

\begin{figure}[htbp]
    \centering
    \includegraphics[width=8cm]{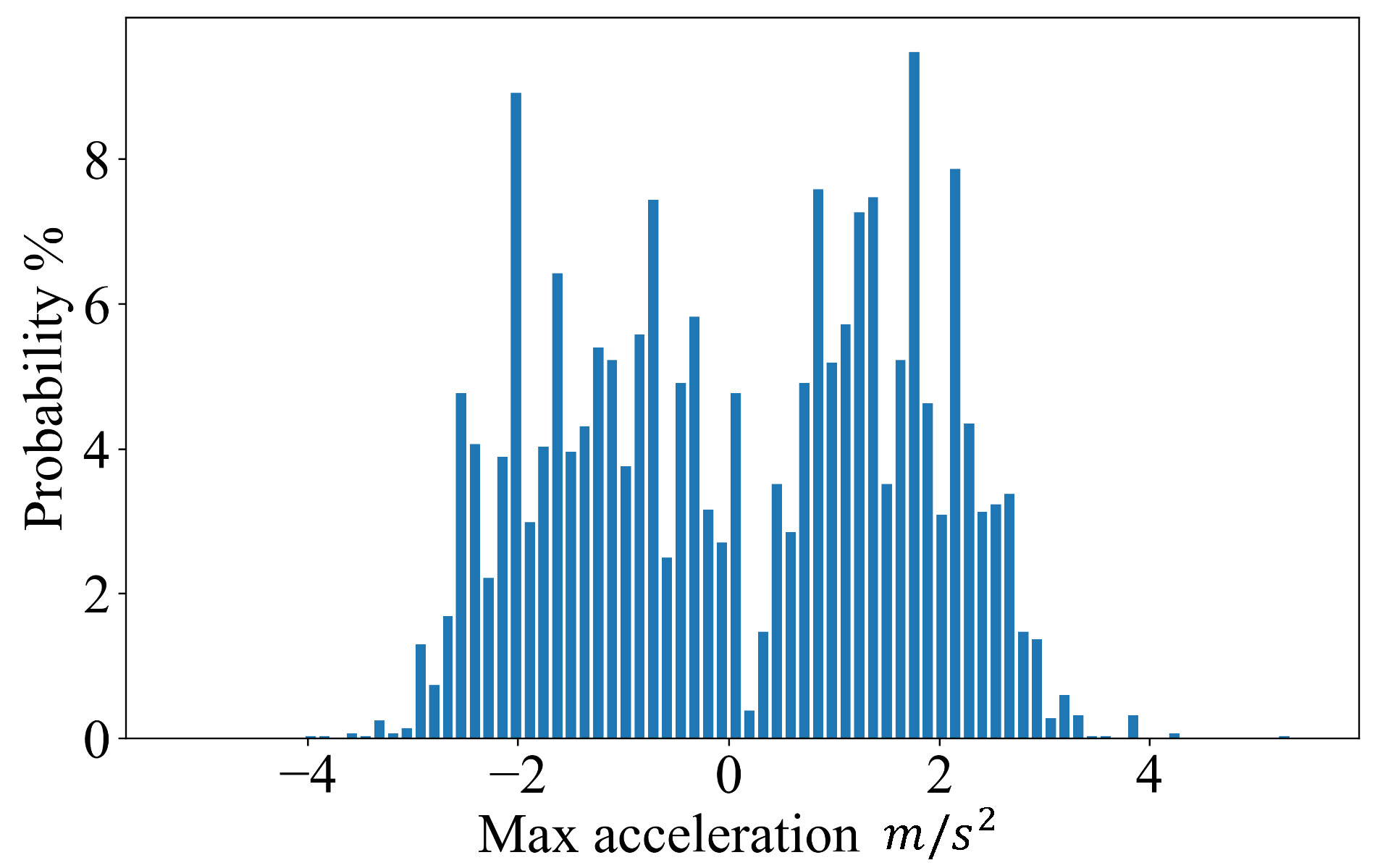}
\caption{The distribution of ground truth max acceleration. }
\label{MaxAcc}
\end{figure}

Kinematic feasibility: The acceleration and  angular velocity is considered for kinematic feasibility. The distribution of maximum acceleration for all the ground truth trajectories is shown in Fig \ref{MaxAcc}. We monitor the absolute value of max acceleration for the generated trajectories and raise an Acceleration Failure (AF) when it exceeds the range of $4m/s^2$. In addition, KL (Kullback-Leibler) divergence of the angular velocity distribution between the generated trajectory data with the log trajectory data is given. The smaller the KL divergence is, the closer it is to the angle velocity distribution of real-world data.

Common sense feasibility: we measure the trajectory collision rate (TCR) as follows.
\[
\rm{TCR} = \frac {\textit{Num}(\rm{collision \ trajectories)}} {\textit{Num}(\rm{all\ generated \  trajectories)}}
\]

\subsubsection{Diversity} 
Following \cite{suo2021trafficsim}, we use a map-aware average self distance (MASD) metric to measure the diversity of the sampled scenarios. In particular, we measure the average distance between the two most distinct sampled trajectories that do not violate traffic rules for each agent in each scenario. 
\begin{equation}
 {\rm{MASD}} = \max_{k,k'\in 1,...,l} \frac{1}{NT_{pred}}\sum_{n=1}^N\sum_{t=1}^{T_{pred}}  \Vert z_{t}^{n,k}-z_{t}^{n,k'} \Vert^2 
\end{equation}

\subsection{Model Details}
LaneGCN model: The input of LaneGCN model contains the HD Map, and the  2\textit{s} history trajectories of agents. The output are the $K=6$ multi-modal trajectories. 
To calculate RouteLoss, we first sample $T_{seq}=10$ waypoints with the equal time interval including startpoint and endpoint from the predicted trajectory. The waypoints and the projected route points from DFS-searched nearest route are used to compute $L_{route}$. The hyper-parameters  $\alpha = 1$ and $\beta = 0.05$ in Eq.(4). The learning rate is set to 1e-3 and the batch size is 16.

TD3 model with attention: 
The policy network consists of three modules of encoder, attention and decoder. The encoder module has three encoders for three kinds of observation $o_r$, $o_e$ and $o_j, j=1,...,5$, each of them contains two dense layers with the same shape of 64, which takes observation as input and outputs features. These features are then fed into the attention module, which contains three nonlinear projections $P_q$, $P_k$, $P_v$ with the same shape of 64. The \textit{attention values} are computed by Eq.(11), and the decoder module processes the \textit{attention values} through two dense layers with the same shape of 256, obtaining the action value. 
The Q-value approximation networks hold the same structure with each other. Each of them contains two modules of encoder and decoder, which has a lighter structure of layers and units compared with the policy network. The observation of the ego and social vehicles are concatenated together to be encoded as a whole rather than encoded separately by different encoders, and the action value as a part of the network's input is concatenated with the encoded state features.
All layers mentioned above are activated by the tanh function.
The learning rate of actor's network is set to 1e-4, and the learning rate of critic's networks is set to 5e-5. The update frequency $d$ is 2 and the batch size is 256.

For the longitudinal and lateral PID controllers, the parameters are designed as follows:
\begin{align*}
\begin{split}
\left \{
\begin{array}{lr}
    {\rm{PID}_{acc}}: k_p=1.0,k_i=0,k_d=0.05\\
    {\rm{PID}_{angle}}:  k_p=1.4,k_i=0.05,k_d=0.25 
\end{array}
\right.
\end{split}
\end{align*}

\begin{figure}[htbp]
    \centering
    \includegraphics[width=8.5cm]{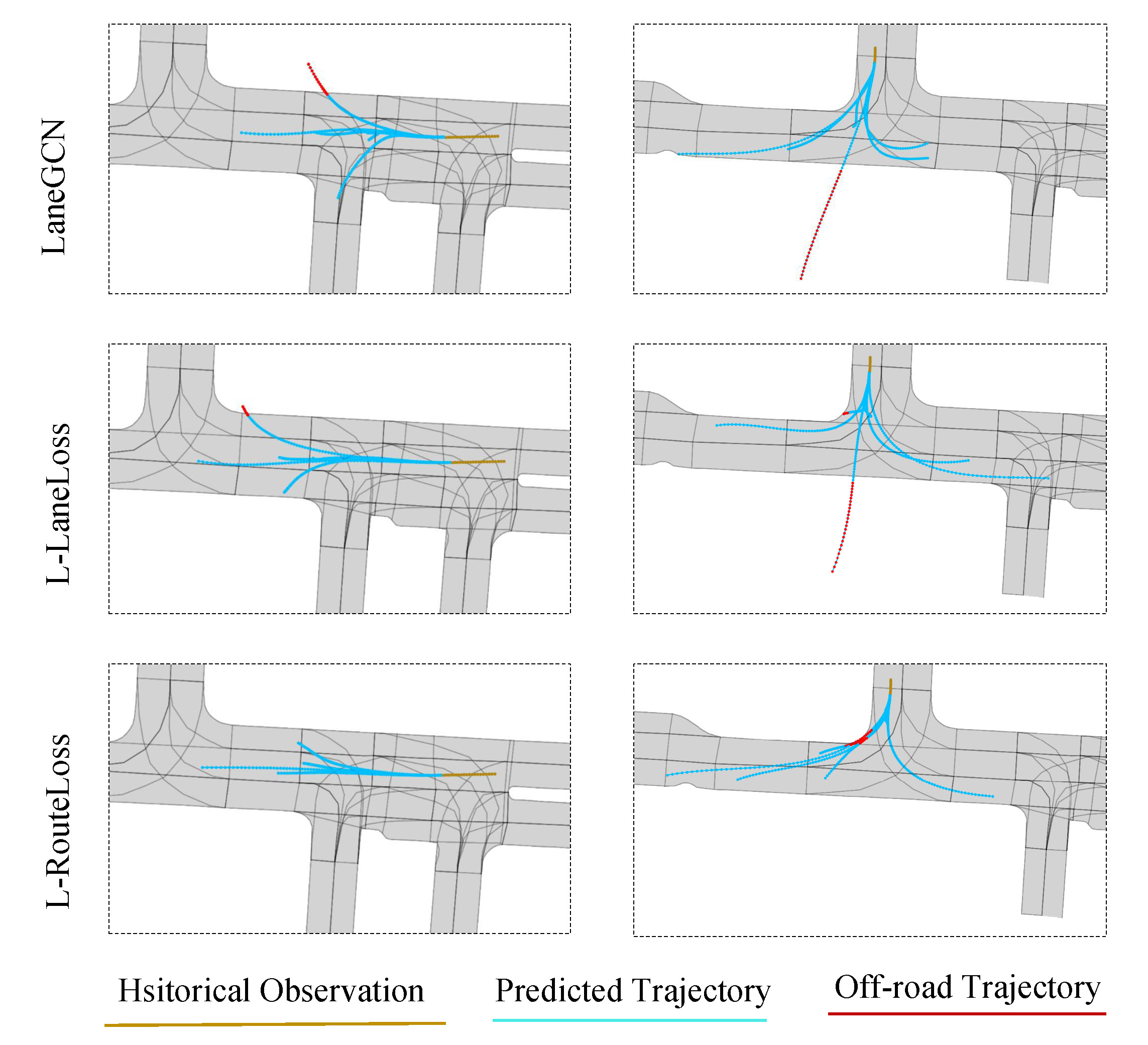}
\caption{Qualitative results for $T=8s$ on typical intersection cases. Lane boundary is shown in black, agent’s past trajectory in yellow, predicted multi-modal trajectories in light blue, and the off-road part in red.}
\label{Laneloss}
\end{figure}

\subsection{Comparison Results}
Here we provide a quantitative evaluation of our proposed trajectory generation method. We generally aim to answer the following three questions.

Q1: Trajectory prediction. Is the auxiliary RouteLoss effective at reducing the off-road trajectories in stage 1?

Q2: Trajectory modification. Can collision behaviors be reduced by the RL model in stage 2 and the kinematic constraints be satisfied with I-Sim?

 Q3: Multi-agent behaviors. Is the proposed TrajGen suitable for multi-agent behaviors?

To answer Q1, we implement the LaneGCN \cite{liang2020learning}, LaneGCN with LaneLoss \cite{anonymous2021improving}, and LaneGCN with RouteLoss proposed in this paper on INTERACTION validation set for the prediction horizon $T=3s$ and $T=8s$. For simplicity, we use L-LaneLoss and L-RouteLoss represent the LaneGCN with LouteLoss and the LaneGCN with RouteLoss, respectively. 
As shown in Table \ref{prediction}, the L-RouteLoss achieves better performances for $T=3s$ in most metrics, especially OR is significantly decreased, which validates the effectiveness of the designed auxiliary RouteLoss. For $T=8s$, its performances are still better in most metrics.  Qualitative results for typical intersection testing cases are shown in Fig. \ref{Laneloss}. It can be seen that the off-road trajectories are reduced with the auxiliary Routeloss. However, OR of L-LaneLoss and L-RouteLoss are close. It may be the off-road trajectory with an endpoint in the lane shown in Fig. \ref{LaneGCNPromblem}(c) is rare for the long horizon prediction.

\begin{table}[h]
\caption{Comparison results of the proposed model on INTERACTION validation set}
\label{prediction}
\begin{center}
\setlength\tabcolsep{1pt}
\begin{tabular}{c|c|c c c c c}
\hline
{$T$} & Model & meanADE$_6$ & meanFDE$_6$ & minADE$_6$ & minFDE$_6$ & OR(\%)\\
\hline
\rule{0pt}{10pt}
\multirow{3}{*}{3s} 
& LaneGCN \cite{liang2020learning} & 1.42&  3.92 & 0.75 & 1.87 &2.05  \\
& L-LaneLoss\cite{anonymous2021improving} & 1.10 & \textbf{3.07} & 0.32 &0.70 &2.13 \\
& L-RouteLoss(ours) & \textbf{1.08} & 3.09 &\textbf{0.30} &\textbf{0.69} &\textbf{1.64} \\[2pt]
\hline
\rule{0pt}{10pt}
\multirow{4}{*}{8s} & LaneGCN \cite{liang2020learning} & 4.65 & 14.03 &  1.96 & 4.20 & 10.82\\
& L-LaneLoss\cite{anonymous2021improving} & 4.50 & 13.40 & 1.43 & 2.85 &\textbf{7.69}\\
& L-RouteLoss(ours) & \textbf{4.38} & \textbf{13.27} & \textbf{1.41} & \textbf{2.83} & 7.88\\[3pt]
\hline
\end{tabular}
\end{center}
\end{table}

\begin{figure}[htbp]
    \centering
    \includegraphics[width=8.5cm]{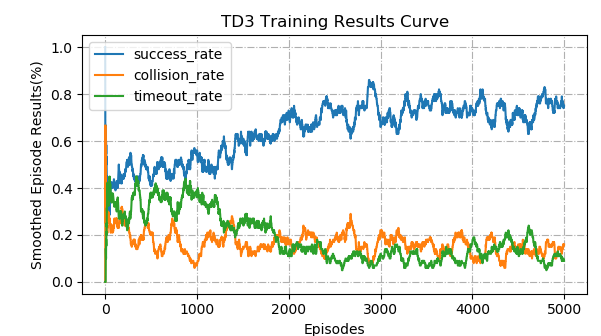}
\caption{RL model's training curve of episode results.}
\label{figure_result}
\end{figure}

To answer Q2, a comparative experiment is conducted between scenarios generated by L-RouteLoss and by TrajGen. 
The RL model of TrajGen is trained on 2849 trajectories from 425 scenarios generated by L-RouteLoss. We randomly select one trajectory at each training episode to be revised using RL model. The episode's result is "collision" if RL controlled vehicle collides with other traffic participants. If RL controlled vehicle reaches trajectory's endpoint within the 10s time horizon, the episode's result is "success", otherwise, the episode's result is "timeout".
The RL model's training curve of episode results can be seen in Fig. \ref{figure_result}, which shows that the RL model achieves a relatively high success rate and low collision and timeout rate within 5000 episodes of training. Then, the trained RL model will be used to modify those infeasible trajectories generated by the predictor.

\begin{figure*}[htbp]
    \centering
    \includegraphics[width=18cm]{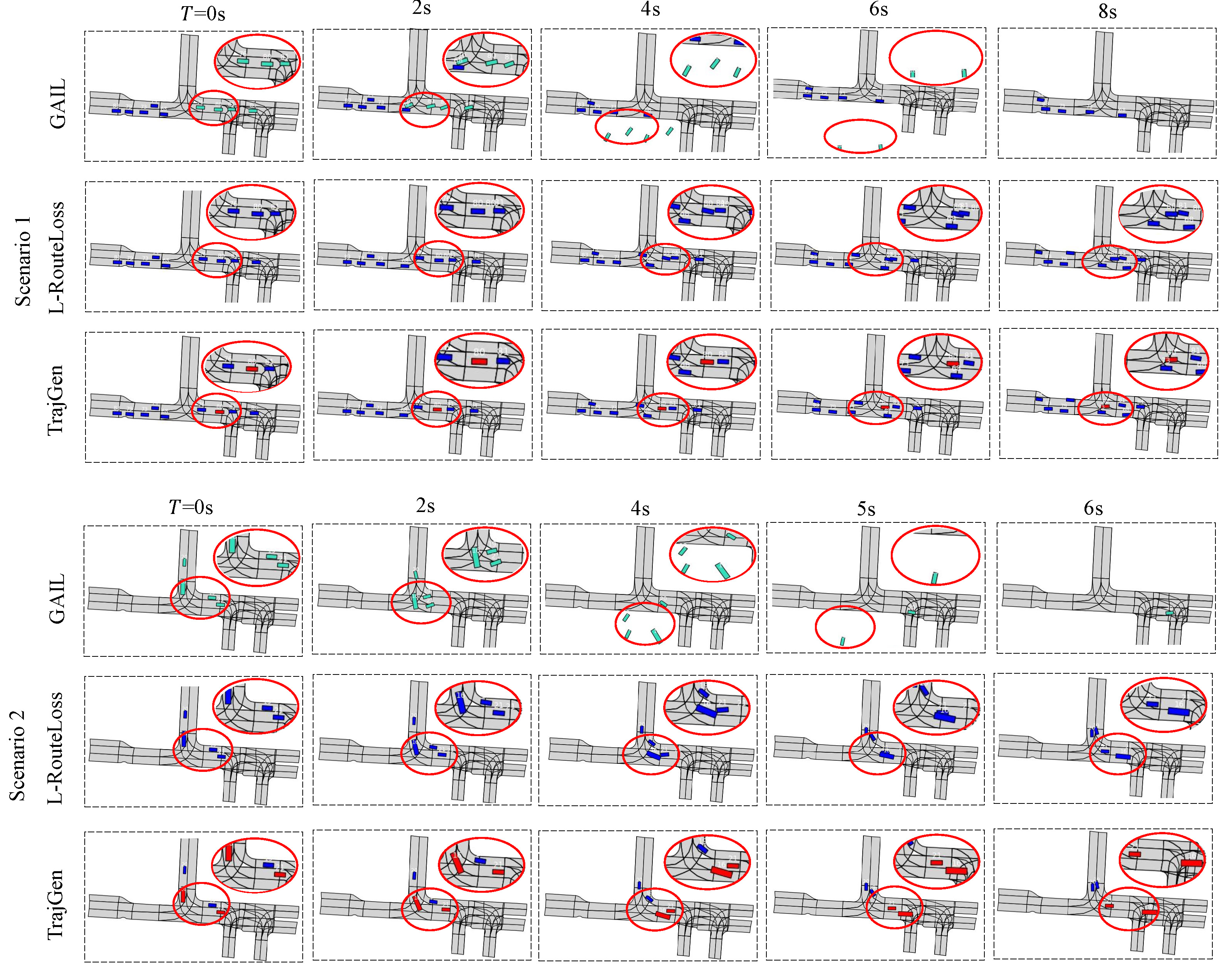}
\caption{Simulated traffic scenarios with sampled key timestamps from trajectory generator GAIL\cite{ho2016generative}, L-RouteLoss and TrajGen. The cyan, blue and red vehicles are controlled by GAIL, L-RouteLoss, and TrajGen, respectively. Since the interactive times of focused vehicles are not the same for scenario 1 and 2, the sampled key timestamps are also different.}
\label{Results}
\end{figure*}

Among the 2849 trajectories generated by L-RouteLoss, there are  828 collision trajectories and 19 kinematic unsatisfied trajectories. Excluding 4 repeated trajectories, there are 843 testing trajectories that need to be modified. Then,
the trained RL model is used to control one collision or kinematic unsatisfied agent to figure whether the collision or kinematic unsatisfied can be avoided.
As shown in Table \ref{TrajGen result}, TrajGen achieves lower RMSE compared with the trajectory prediction model L-RouteLoss, which means the higher matching degree and fidelity with ground truth trajectory. 
For the reactivity, 233 synthetic scenarios' test shows that TrajGen's collision rate decreases 15.8\% compared with L-RouteLoss. 
For the feasibility, with the help of RL model, collision trajectories are reduced from 828 to 435, thus TCR is also reduced from 28.9\% to 15.3\%, which means nearly half of the collision trajectories are reduced as shown in Scenario 1 of Fig. \ref{Results}. 
Meanwhile, it shows that the AF and KL divergence of TrajGen, which considers the bicycle model during the training process, is significantly reduced compared with the L-RouteLoss prediction model.
In addition, the MASD of L-RouteLoss and TrajGen are very close, which means TrajGen can maintain the diversity from the first multi-modal prediction stage.

For comparison, we also train a GAIL model \cite{ho2016generative} as the trajectory generator. In general, we should first focus on fidelity for simulation scenarios, where the RMSE of GAIL is too poor to be acceptable. The reason is that GAIL cannot keep the vehicle driving in the lane as shown in Fig. \ref{Results}. As the vehicle controlled by GAIL can not drive along the lane well, it may have moved out of the drivable area before colliding with a static car in synthetic scenarios. This is the reason why the SCR of GAIL is lower than TrajGen. In addition, the MASD of GAIL is zero since it can not generate multi-modal trajectories. In other words, GAIL is almost impossible to generate high-quality trajectories with  a long horizon. Overall, the proposed TrajGen outperforms the trajectory prediction method and GAIL in terms of comprehensive metrics.


\begin{table}[h]
\caption{Comparison quality of generated simulation scenarios}
\label{TrajGen result}
\setlength\tabcolsep{3pt}
\renewcommand{\arraystretch}{1.3}
\begin{tabular}{c|c|c|ccc|c}
\hline
\multirow{2}{*}{Model} & Fidelity & Reactivity & \multicolumn{3}{|c|}{Feasibility}                                 & Diversity \\ 
\cline {2 - 7} & \text{RMSE}$\downarrow$  & SCR$\downarrow$  & \multicolumn{1}{c}{TCR$\downarrow$} & \multicolumn{1}{c}{AF$\downarrow$} & KL$\downarrow$ & MASD$\uparrow$  \\ \hline
Log replay             & /        & 100\%        & \multicolumn{1}{c}{/ }       & \multicolumn{1}{c}{/ }  &  /    & /         \\ \hline
GAIL {[}32{]}          & 20.96    &   \textbf{19.9\%}         & \multicolumn{1}{c}{45.8\%}        & \multicolumn{1}{c}{\textbf{0}}   &  3.58       & 0         \\ \hline
L-RouteLoss(ours)            & 4.21     & 41.6\%       & \multicolumn{1}{c}{28.9\%}    & \multicolumn{1}{c}{19} & 3.44   & \textbf{6.71}      \\ \hline
TrajGen(ours)          & \textbf{3.58}     & 25.8\%       & \multicolumn{1}{c}{\textbf{15.3\%}}    & \multicolumn{1}{c}{\textbf{0}}  & \textbf{1.05}     & 6.56      \\ \hline
\end{tabular}
\end{table}


To answer Q3, we deploy the TrajGen model on the two collision agents using the parameters sharing technique. A typical scenario can be seen in Scenario 2 of Fig. \ref{Results}, where the predicted collision trajectories can not be reduced if we choose either agent to deploy TrajGen. The results show that the collision can be avoided by depolying TrajGen on those two agents, and the generated scenarios are naturalistic. However, for the GAIL model with parameter sharing, the multiple controlled vehicles are out of drivable area which generates an invalid scenario. Therefore, the proposed TrajGen is also suitable for multi-agent behaviors based on parameters sharing.


\section{Conclusion}
In this paper, we have proposed a novel framework to generate realistic and diverse trajectories with a reactive and feasible agent behavior model based on naturalistic driving data. TrajGen is a two-stage trajectory generation model for background vehicles in simulation scenarios, which is trained directly from real-world INTERACTION dataset. In addition, we develop a data-driven simulator I-Sim that can be used for RL training based on real-world traffic scenarios. The experimental results show that TrajGen can reduce nearly half of the collision trajectories, and the quality of generated trajectories is improved obviously in  terms  of fidelity, reactivity, feasibility, and  diversity. 

Scenario generation with data-driven behavior model is still a challenging and important task for autonomous driving now. In future work, we will further exploit the generalization ability of TrajGen in different kinds of complicated scenarios. In addition, evaluating the existing planning models and finding their corner case scenarios based on the scenario generation approach is another interesting and important work.

\bibliographystyle{IEEEtran}

\bibliography{ref} 



\end{document}